\documentclass[letterpaper]{article} 
\usepackage{aaai24}  
\usepackage{times}  
\usepackage{helvet}  
\usepackage{courier}  
\usepackage[hyphens]{url}  
\usepackage{graphicx} 
\usepackage{natbib}  
\usepackage{caption} 
\usepackage{algorithm}
\usepackage{algorithmic}

\usepackage{newfloat}
\usepackage{listings}
\DeclareCaptionStyle{ruled}{labelfont=normalfont,labelsep=colon,strut=off} 
\floatstyle{ruled}
\newfloat{listing}{tb}{lst}{}
\floatname{listing}{Listing}
\usepackage{algorithm}
\usepackage{algorithmic}
\usepackage{subfig}
\usepackage[title]{appendix}
\usepackage{color}

\usepackage{amsmath}
\usepackage{bbold}
\usepackage{mathtools}
\usepackage{xspace}
\usepackage{tabularx}

\newcommand{\etal}{\emph{et~al.}\xspace}
\newcommand{\eg}{\emph{e.g.},\xspace}

\newcommand{\ie}{\emph{i.e.},\xspace}

\newcommand{\eat}[1]{}
\ifodd 2
\newcommand{\TODO}[1]{{\color{red}TODO:{#1}}}
\newcommand\beftext[1]{{\color[rgb]{0.5,0.5,0.5}{BEFORE:#1}}}
\else
\newcommand{\TODO}[1]{}
\newcommand{\beftext}[1]{}
\fi
\newcommand*\samethanks[1][\value{footnote}]{\footnotemark[#1]}
\usepackage{booktabs}

\title{A Cross-View Hierarchical Graph Learning Hypernetwork for Skill Demand-Supply Joint Prediction}

\author {
    Wenshuo Chao\textsuperscript{\rm 1,\rm 2},
    Zhaopeng Qiu\textsuperscript{\rm 2},
    Likang Wu\textsuperscript{\rm 2,\rm 3},
    Zhuoning Guo\textsuperscript{\rm 1}, \\
    Zhi Zheng\textsuperscript{\rm 2,\rm 3},
    Hengshu Zhu\textsuperscript{\rm 2,\rm 1}\thanks{Corresponding Author},
    Hao Liu\textsuperscript{\rm 1}\samethanks
}

\affiliations {
    \textsuperscript{\rm 1} The Hong Kong University of Science and Technology (Guangzhou) \\
    \textsuperscript{\rm 2} Career Science Lab, BOSS Zhipin \\
    \textsuperscript{\rm 3} University of Science and Technology of China \\
    \{wchao829,zguo772\}@connect.hkust-gz.edu.cn, 
    \{zhpengqiu, zhuhengshu\}@gmail.com, 
    \{wulk,zhengzhi97\}@mail.ustc.edu.cn,
    liuh@ust.hk
}

\usepackage{bibentry}

\begin{document}

\maketitle

\begin{abstract}
\beftext{The rapidly evolving landscape of technology and industries leads to changes in skill requirements, posing challenges for individuals to adapt to the dynamic work environment.
Anticipating these skill shifts is vital for employees to choose the right skills to learn and maintain competitive edges in the job market.
However, the existing research rarely adopts the machine learning approach to predict the evolution of skills. Of the few that do, they resort to basic time series forecasting problems. The methods neglect the complicated relationship between different skills and overlook the joint prediction of skill demand and supply, which is crucial for individuals when deciding which skills to acquire.
In this paper, we propose a Cross-view Hierarchical Graph learning with hypernetwork model (CHG) for jointly predicting skill demand-supply to bridge the gap between employees and employers.
In particular, we first propose a cross-view graph encoder to capture the interconnection between supply and demand.
Then, a hierarchical encoder is introduced for modeling the impact of technology and policy changes that affect skills from a cluster-wise perspective. 
Finally, the conditional hyper-decoder is introduced to enhance the capability of joint prediction of supply and demand by incorporating historical gap changes.  
Experiments on three real-world datasets demonstrate the effectiveness of the introduced modules and the superiority of the proposed model compared to state-of-the-art baselines.}

The rapidly changing landscape of technology and industries leads to dynamic skill requirements, making it crucial for employees and employers to anticipate such shifts to maintain a competitive edge in the labor market. Existing efforts in this area either rely on domain-expert knowledge or regarding skill evolution as a simplified time series forecasting problem. However, both approaches overlook the sophisticated relationships among different skills and the inner-connection between skill demand and supply variations. 
In this paper, we propose a Cross-view Hierarchical Graph learning Hypernetwork (CHGH) framework for joint skill demand-supply prediction. Specifically, CHGH is an encoder-decoder network consisting of i) a cross-view graph encoder to capture the interconnection between skill demand and supply, ii) a hierarchical graph encoder to model the co-evolution of skills from a cluster-wise perspective, and iii) a conditional hyper-decoder to jointly predict demand and supply variations by incorporating historical demand-supply gaps. Extensive experiments on three real-world datasets demonstrate the superiority of the proposed framework compared to seven baselines and the effectiveness of the three modules. 

\end{abstract}

\section{Introduction}

\begin{figure}[t]
    \centering
    \includegraphics[width=0.47\textwidth]{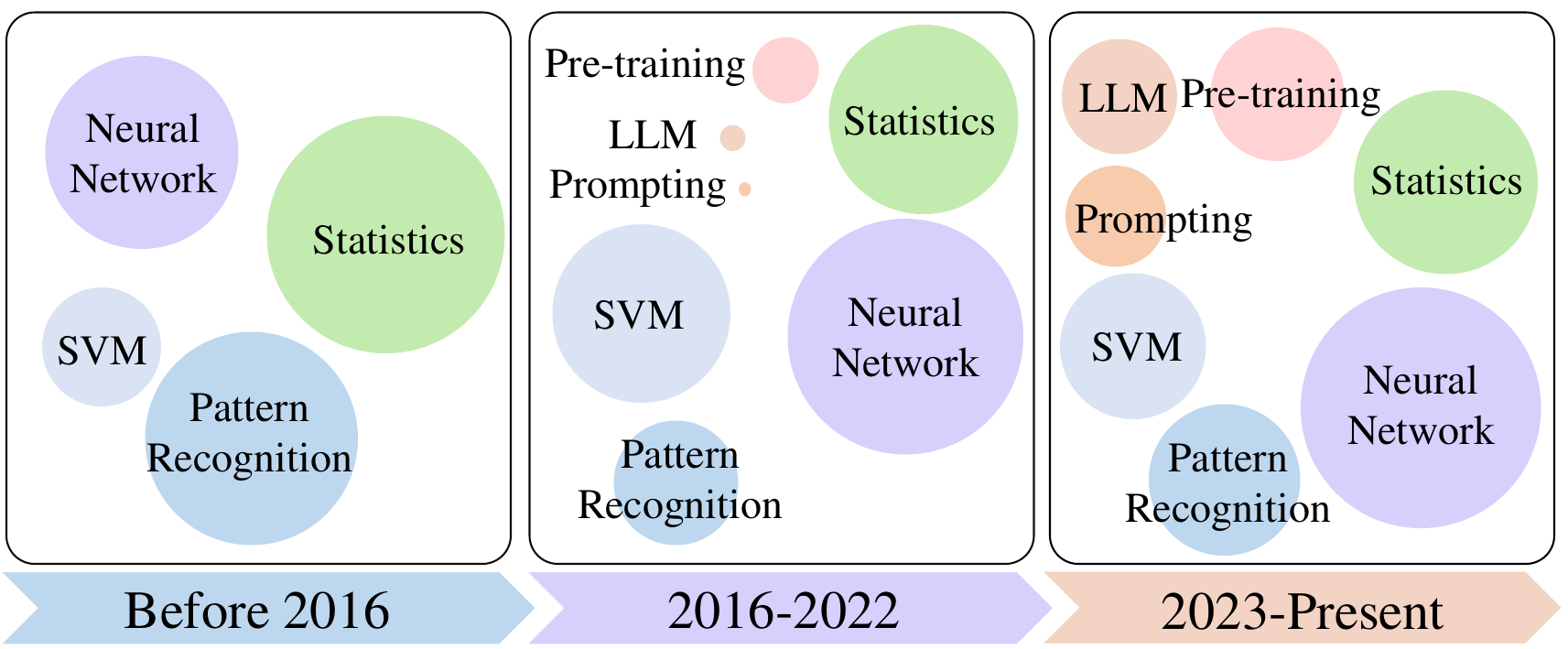}
    \caption{Real-world skill demand variation of machine-learning related jobs, data ranged from 2015 to 2023.}
    \label{fig: Skill gap}
\end{figure}

The rapid advancement of information technology (\eg 5G, VR, and generative AI) has significantly reshaped the skill requirements.
According to the released reports, the required skill set in the job market has changed by about 25\% since 2015~\cite{LinkedInSkill}.
Take machine-learning-related skills for example, there has been a booming demand for deep-learning-related skills since 2014 and Large Language Model (LLM)~\cite{radford2018improving} related skills in the past year, but declining demand on classic statistical learning skills, as illustrated in Figure~\ref{fig: Skill gap}.
Such shifts in skill requirements have led to a significant imbalance of skill demand and supply in various industries, commonly referred to as the skill gap~\cite{larsen2018developing}, which may result in unemployment and business failures as industries evolve~\cite{kim2006update,mcguinness2016skill, donovan2022skills}.
It is crucial to forecast and analyze the skill supply and demand to help various stakeholders anticipate and proactively handle the skill gap. For instance, job seekers can plan relevant skill-learning paths in advance, companies can optimize their workforce planning strategies, and policymakers can develop policies to guide market development.

Existing studies on skill demand and supply prediction can be broadly categorized into expert-based and learning-based methods. 
Expert-based methods analyze skill shifts by exploring survey data~\cite{belhaj2013markov}.
Such methods usually deliver coarse-grained forecasting results via expert domain knowledge~\cite{bughin2018skill}.
Learning-based methods predict more fine-grained skill shifts by exploiting patterns and implications based on machine learning techniques.
For example, Xu \etal~\cite{xu2018measuring} measure the popularity of skills via topic model and Macedo \etal~\cite{garcia2022practical} predict skill demands by adopting the recurrent neural network.
However, both classes of approaches overlook the sophisticated demand-supply relationships among different skills grounded in structural and contextual domains. 
In this work, we investigate the joint prediction of skill demand and supply by explicitly modeling the collaborative relationships between different skills.

However, this is a non-trivial task due to the following technical challenges.
First, the demand and supply variation of each skill in the labor market is correlated yet diversified. For instance, the breakthrough of LLM led to a great demand for prompt engineering, and the supply of this skill may slightly be lagged behind the demand to let talents familiar with it.
Capturing the interconnection between demand and supply variations perhaps can enhance one another predictive tasks.
Second, the demand-supply variation of different skills is not standalone. Similar skills (\eg \textit{C++} and \textit{Java}) may share an evolution trend conditioned on the macro-level technology advancement and economic status.
The challenge is identifying and preserving the synchronized evolution patterns of different skills to improve overall prediction accuracy.
Third, despite the joint modeling paradigm sharing knowledge between skill demand and supply variations, the demand and supply predictions are still made separately. 
How to collectively output skill demand-supply predictions to further calibrate supply and demand predictions is another challenge.

In this paper, we propose a \textbf{C}ross-View \textbf{H}ierarchical \textbf{G}raph Learning \textbf{H}ypernetwork (CHGH) framework to predict skill supply and demand variations jointly. 
Specifically, the CHGH framework consists of a cross-view graph encoder module that enhances skill demand and supply representations by capturing their asymmetric relationship through an inter-view adaptive matrix.
Besides, we devise a hierarchical graph encoder that models the high-level skill co-evolve trends by harnessing the cluster-wise skill variation correlation.
Furthermore, we propose a hyper-decoder that outputs skill variations conditioned on the historical skill demand-supply gaps. By incorporating the paired relationship between the supply and demand of each skill, the hyper-decoder derives more accurate predictions to better support downstream analysis.
The contributions of the paper are summarized as follows:

\begin{itemize}
    \item To our knowledge, this is the first work that investigates the skill demand-supply joint prediction problem. By anticipating future skill demand and supply variations, employees and employers can effectively identify potential skill gaps and address them accordingly. 
    \item We propose a cross-view hierarchical graph learning with hyper-decoder framework that simultaneously captures inter-dependencies between the skill supply and demand views, as well as the asymmetric cluster-wise correlation between skills.
    \item We evaluate the effectiveness of the proposed approach on three real-world datasets, and the experimental demonstrates the superiority of CHGH compared with seven baselines.
\end{itemize}    

\section{Preliminaries}

\newtheorem{definition}{Definition}

In this section, we first introduce the real-world datasets used in this paper. Then, we formally define key concepts used in our task. Finally, we present the problem formulation of skill demand-supply joint prediction.

\subsection{Data Description}\label{sec:data}
The real-world dataset used in this work is extracted from an online recruitment platform, consisting of 2,254,733 job postings and 3,545,908 work experience descriptions.
We divide the datasets into three main categories following the Global Industry Classification Standard (GICS)~\cite{phillips2016industry}, including Information Technology (IT), Finance (FIN), Consumer Discretionary and Consumer Staples (CONS), over the period from September 2017 to March 2019. 
The details of the dataset are illustrated in Table~\ref{dataset}. 
In identifying job skills, we combined a list from Emsi Burning Glass~\cite{garcia2022practical} with our own additional skills, ensuring a comprehensive mix that addresses current industry trends.
We preserve 7,446 skills that exist in the dataset by removing the rest that are too sparse in the job description. 
The occurrence relationship of skills in the job description is reported in Figure~\ref{fig:datadescription:a}.

\begin{table}[t]
\centering
\begin{tabular}{c|ccc}
\hline 
Datasets & IT & FIN & CONS \\
\hline 
\# of job description & 902,442 & 645,899 & 692,163 \\
\hline 
\# of work experience & 363,376 & 223,330 & 455,120 \\

\hline
\end{tabular}
\caption{Statistics of datasets.}
\label{dataset}
\end{table}

\subsection{Skill Supply and Demand Quantification}

To quantify skill supply and demand shift, we first denote the skill set as $\mathcal{K}$ and discretize the investigated period into equal time steps $\mathcal{T}$. 
The job description $jd$ and work experience $we$ sets located at time step $t \in \mathcal{T}$ are denoted as $\mathcal{J}_t$ and $\mathcal{W}_t$, respectively.
Then, we define the \emph{Demand share} and \emph{Supply share} to measure the relative importance of skills skill demand and supply,
which indicates the proportion of a particular skill that appeared in the job description and work experience~\cite{garcia2022practical}. 

\begin{definition}{\textbf{Demand share}.} 
    Demand share $ \mathcal{D}_k^t$ is defined as the percentage of job descriptions that required the skill $k \in \mathcal{K}$ at the given time step $t \in \mathcal{T}$. 
    The indicator function $\mathbb{1}_{k \in jd}$ is used to represent whether the skill $k$ appears in the job description.
    $|\mathcal{J}_t|$ represents the total number of job descriptions at time step $t$.
    \begin{equation}
        \mathcal{D}_k^t =  \frac{\Sigma_{{jd} \in \mathcal{J}_t}{\mathbb{1}_{k \in jd}}}{|\mathcal{J}_t|}.
    \end{equation}
\end{definition}

\begin{definition}{\textbf{Supply share}.}
    Supply share $ \mathcal{S}_k^t$ is defined as the percentage of work experience that contained the skill $k \in \mathcal{K}$ at the time step $t \in \mathcal{T}$. 
    $|\mathcal{W}_t|$ represents the total number of work experience at time step $t$.
    \begin{equation}
        \mathcal{S}_k^t = \frac{\Sigma_{{we} \in \mathcal{W}_t}{\mathbb{1}_{k \in we}}}{|\mathcal{W}_t|}.
    \end{equation}
\end{definition}

\begin{figure}[t]
    \label{fig:datadescription}
    \centering
    \subfloat[Occurrence of skills in job descriptions. \label{fig:datadescription:a}]{\includegraphics[width=0.20\textwidth]{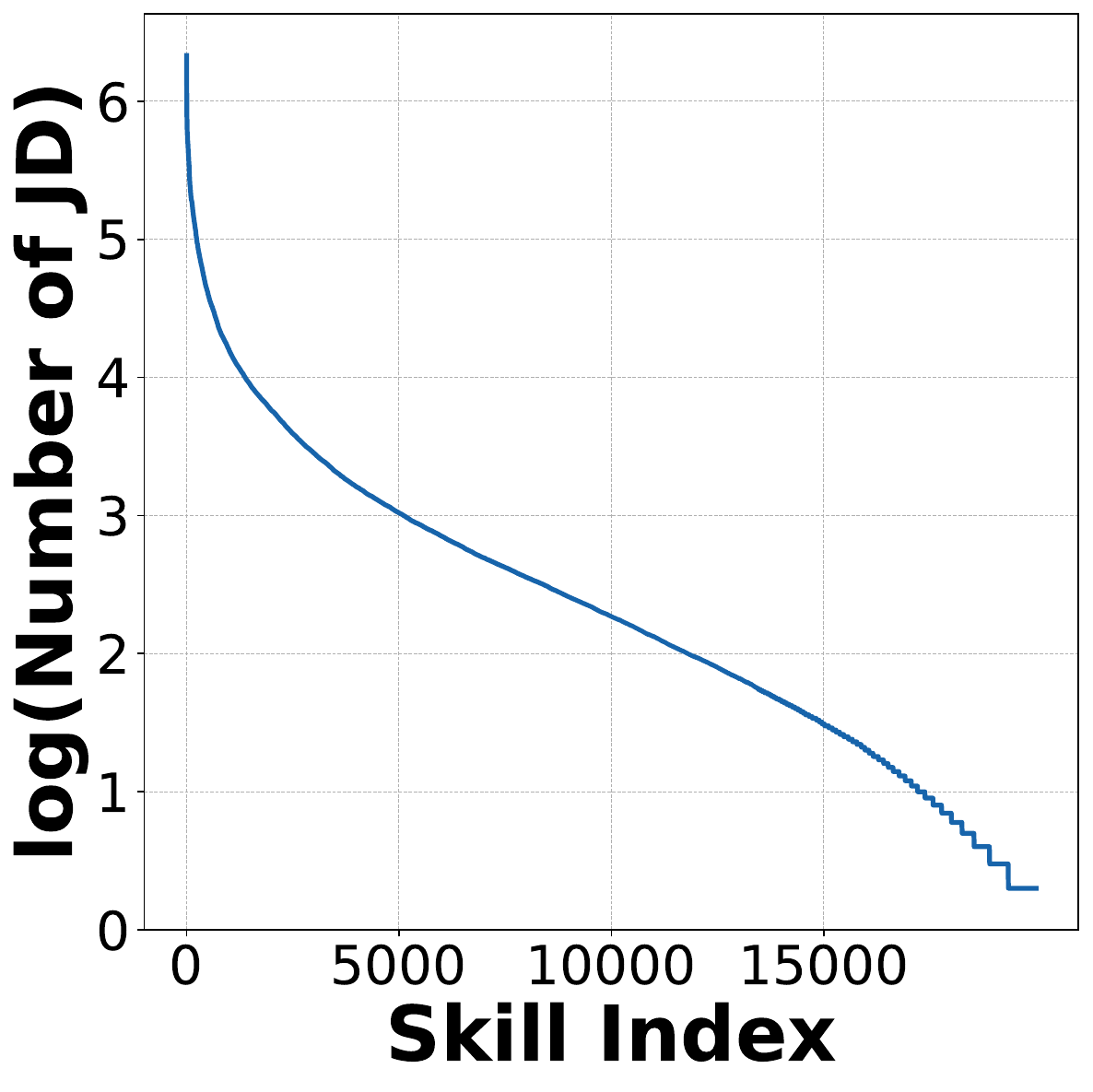}}
    \hfill
    \subfloat[Demand-supply distribution of skills. \label{fig:datadescription:c}]{\includegraphics[width=0.21\textwidth]{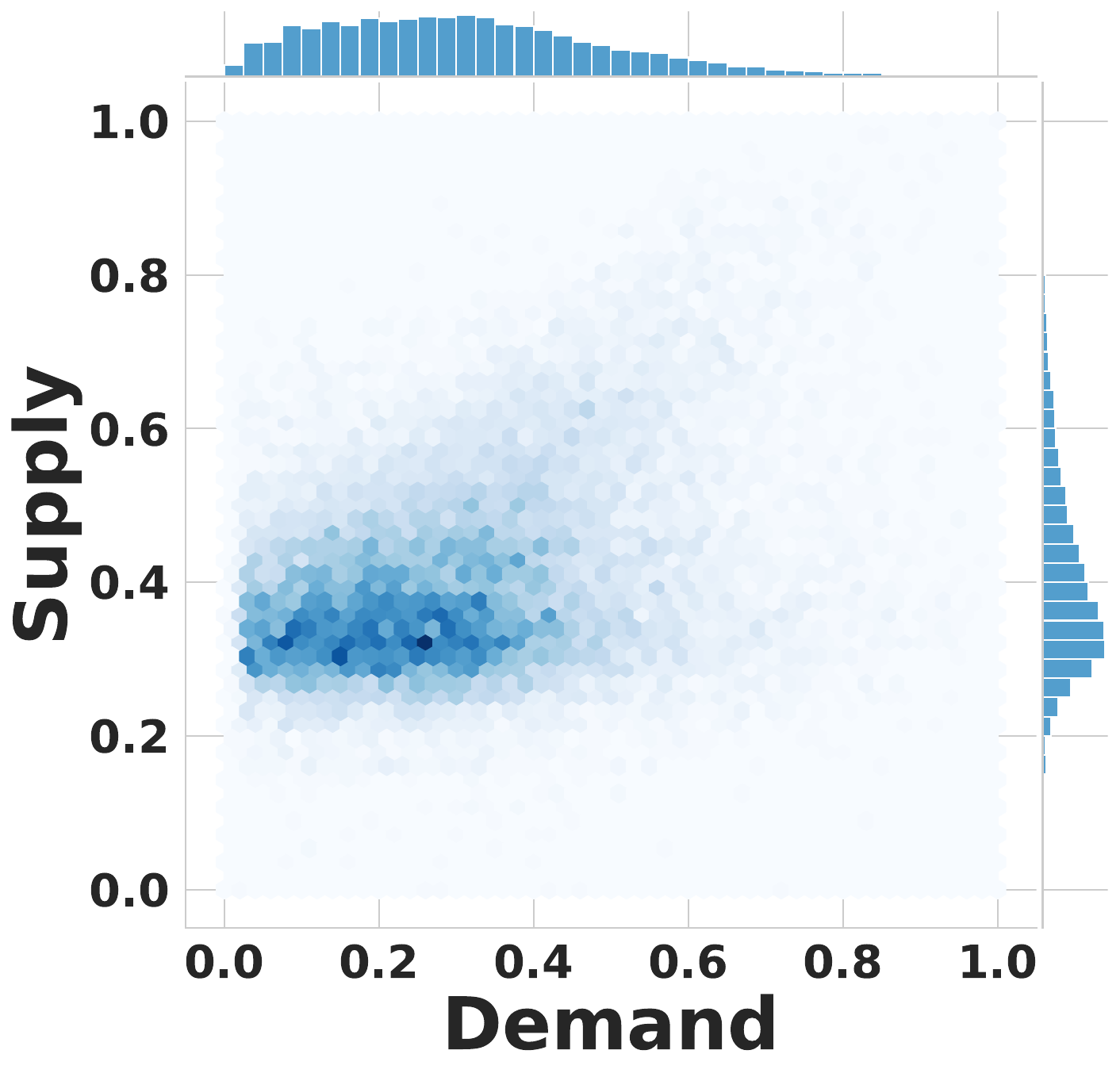}}
    \caption{Skill distribution on IT dataset.}
\end{figure}

In this work, we use one month as the basic time step to derive demand share and supply share.
Figure~\ref{fig:datadescription:c} depicts the distribution of skill demand-supply relationships. We can observe a strong positive correlation between skill demand and supply, which validate the necessity of joint demand-supply prediction. We further quantify the demand share and supply share difference. 

\begin{definition}{\textbf{Skill gap}.}
The skill gap $SG_k^t$ for skill $k$ at the time step $t$ is defined as $SG_k^t = \mathcal{D}_k^t - \mathcal{S}_k^t$. 
\label{def:skillgap}
\end{definition}

The skill gap represents the demand and supply imbalance in the labor market, indicating the scarcity or abundance of skill $k$. 
Then we quantify the correlation between skills based on the co-occurrence information.

\begin{definition}{\textbf{Skill demand graph}.}
Skill demand relation graph is defined as $\mathcal{G}_\mathcal{D} = (\mathcal{V},\mathcal{E}_\mathcal{D},\mathbf{A}^\mathcal{D})$. This graph represents the co-occurrence of skills in all job descriptions $ \mathcal{J} $ of the training data. $\mathcal{V} = \mathcal{K}$ is the set of skills, and $\mathcal{E}_\mathcal{D}$ contains edges with weights that signify the normalized co-occurrence of skills in $\mathcal{J}$. 
$\mathbf{A}^\mathcal{D}_{i,j}$ is the corresponding weighted adjacency matrix
\begin{equation}
\mathbf{A}^\mathcal{D}_{i,j} = 
\begin{cases}
\begin{split}
&R_{i,j,\mathcal{J}}, && \text{if} \quad R_{i,j,\mathcal{J}}> \epsilon  \\
&0, && \text{otherwise} \\
\end{split}
\end{cases},
\end{equation}
\begin{equation}
\label{eq:normalized-cooc}
R_{i,j,\mathcal{J}} = \frac{\Sigma_{{jd} \in \mathcal{J}}{\mathbb{1}_{k_i,k_j \in jd} }}{\Sigma_{{jd} \in \mathcal{J}}{\mathbb{1}_{k_i \in jd}}},
\end{equation}
where $R_{i,j,T}$ is the normalized co-occurrence ratio of skill $k_i$ and $k_j$, and $\epsilon$  is a occurrence threshold.
\end{definition}

\begin{definition}{\textbf{Skill supply graph}.}
Skill supply relation graph is defined as $\mathcal{G}_\mathcal{S} = (\mathcal{V},\mathcal{E}_\mathcal{S},\mathbf{A}^\mathcal{S})$. This graph represents the co-occurrence of skills in all work experience descriptions $\mathcal{W}$ of the training data. 
It is constructed in a similar way as $\mathcal{G}_\mathcal{D}$, by utilizing $R_{i,j}^{\mathcal{W}}$ as normalized co-occurrence ratio in Eq.~(\ref{eq:normalized-cooc}).
$\mathbf{A}^\mathcal{S}_{i,j}$ is the weighted adjacency matrix
\begin{equation}
\mathbf{A}^\mathcal{S}_{i,j} = 
\begin{cases}
\begin{aligned}
&R_{i,j,\mathcal{W}}, && \text{if} \quad R_{i,j,\mathcal{W}}> \epsilon  \\
&0, && \text{otherwise} \\
\end{aligned}
\end{cases}.
\end{equation}
\end{definition}

\subsection{Problem Statement}
Given the demand sequence of all skills
$\mathcal{D}^\mathcal{T}_\mathcal{K} = \{\mathcal{D}_{\mathcal{K}}^0, \mathcal{D}_{\mathcal{K}}^1,...\mathcal{D}_{\mathcal{K}}^t\}$, where $\mathcal{D}_{\mathcal{K}}^t \in \mathbb{R}^{|\mathcal{K}|\times 1}$ denotes the demand share at time step $t$, the
supply sequence 
$\mathcal{S}^\mathcal{T}_\mathcal{K}$, 
and skill gap historical sequence in time series 
$SG^\mathcal{T}_\mathcal{K} = \{SG_{\mathcal{K}}^0,SG_{\mathcal{K}}^1,...SG_{\mathcal{K}}^t\}$ with elements $SG_{\mathcal{K}}^t \in \mathbb{R}^{|\mathcal{K}| \times 1}$. Our task is to utilize the skill co-occurrence graphs from the supply ($\mathcal{G}_\mathcal{S}$) and demand ($ \mathcal{G}_\mathcal{D}$) views to jointly predict the demand and supply share of each skill $k$ in the next time step,
\begin{equation}
\hat{Y}^{t+1}_{\mathcal{D}},\hat{Y}^{t+1}_{\mathcal{S}} \gets  \mathcal{F}(\mathcal{S}^{\mathcal{T}}_{\mathcal{K}}, \mathcal{D}^{\mathcal{T}}_{\mathcal{K}}, SG^{\mathcal{T}}_{\mathcal{K}},\mathcal{G}_\mathcal{S}, \mathcal{G}_\mathcal{D}), 
\end{equation}
where $\hat{Y}^{t+1}_{\mathcal{D}},\hat{Y}^{t+1}_{\mathcal{S}}$ are the estimated supply and demand share of all skills at time step $t+1$.

\section{Methodology}

\begin{figure*}[t]
    \centering
    \includegraphics[width=\textwidth]{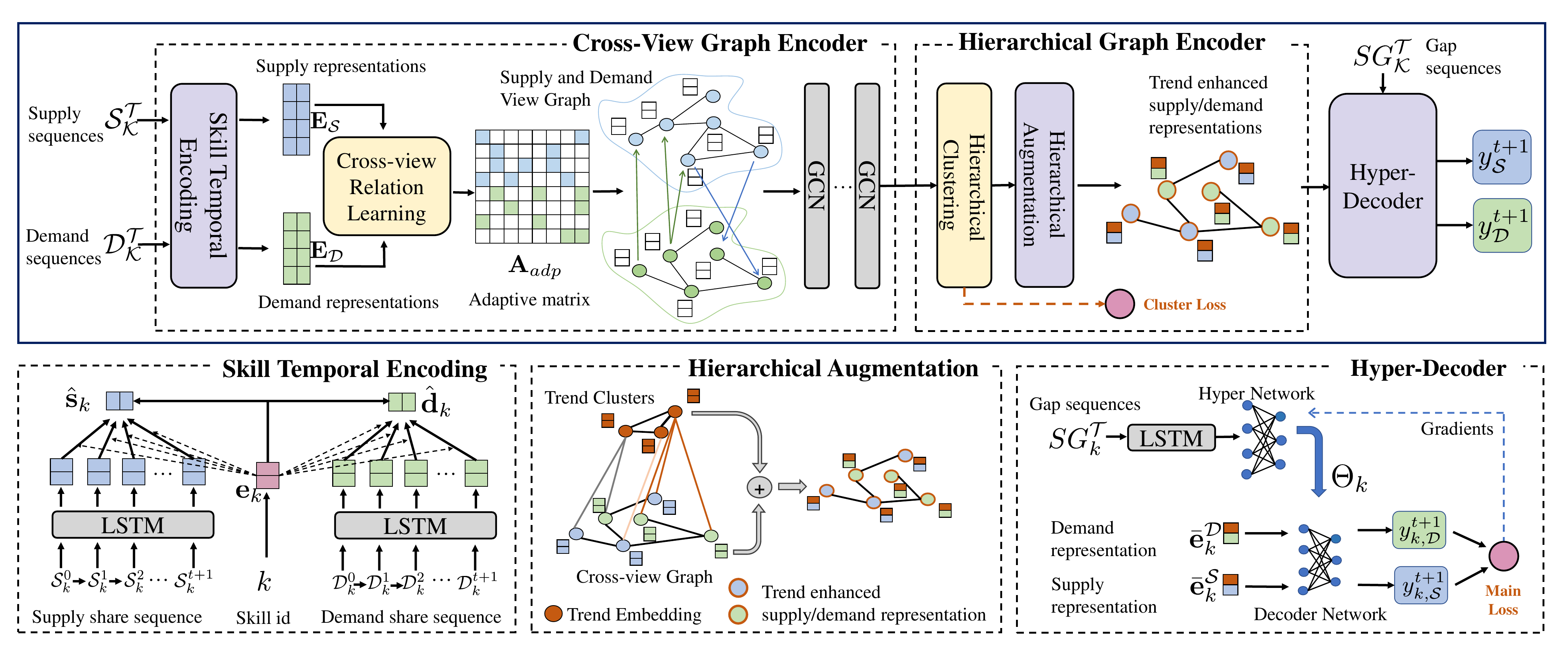}
    \caption{The CHGH framework mainly consists of three modules. 
    (1) The cross-view graph encoder captures the asymmetric correlation for each skill between the supply and demand views.
    (2) The hierarchical graph encoder is responsible for preserving high-level co-evolution patterns between similar skills.
    (3) The conditional hyper-decoder jointly optimizes the supply and demand decoders based on historical skill gap tendencies.}
    \label{fig:framework}
\end{figure*}

In this section, we present the CHGH framework in detail. As depicted in Figure~\ref{fig:framework}, CHGH follows the encoder-decoder architecture comprising three major modules.

\subsection{Cross-view Graph Encoder (CGE)}

Existing studies on skill variation forecasting solely consider the dependencies in either demand or supply view. 
In fact, the demand and supply variation of skills are correlated yet diversified. To share the knowledge between demand and supply graph views~\cite{khan2019multi,liang2022multi}, we propose the cross-view graph encoder, which consists of three components: skill temporal encoding, cross-view relation learning, and cross-view augmentation.
Specifically, a skill temporal encoding block initially captures the temporal dependency within skill supply and demand sequences. After that, a cross-view relation learning block is invoked to identify the intricate asymmetric skill relationships between the supply and demand views.
Finally, we employ a cross-view augmentation block to propagate knowledge from one view to another through the learned cross-view asymmetric skill relationships.

\subsubsection{Skill temporal encoding.}
We first convert each skill $k\in \mathcal{K}$ to a \(d\)-dimensional vector \(\mathbf{e}_k\) via an embedding matrix \(\mathbf{E}\in \mathbb{R}^{|\mathcal{K}|\times d}\), where \(d\) is the dimension of each skill embedding.
Take the supply sequence ${\mathcal{S}}^{\mathcal{T}}_{k} \in \mathbb{R}^{|\mathcal{T}|\times1}$ of skill $k$ for illustration, we further convert it to a matrix ${\mathbf{S}}^{\mathcal{T}}_{k} \in \mathbb{R}^{|\mathcal{T}|\times d}$ via a two-layer MLP.
Then, we use two Long Short Term Memory (LSTM) layers to capture the temporal shifts,

\begin{equation}
\mathbf{h}^t_k = \operatorname{LSTM}(\mathbf{S}^{t}_k,\mathbf{h}^{t-1}_k),
\end{equation}
where $\mathbf{S}_k^{t}$ is the supply embedding derived in time step $t$ and $\mathbf{h}^0_k$ is initialized by the skill embedding $\mathbf{e}_k$. 
After that, we adopt the attention mechanism $\operatorname{Attn}(\cdot)$~\cite{wang2016attention} that adaptively reweighting the essential temporal steps for different skills to obtain the supply sequence-level representation,
\begin{equation}
\label{eq:aggregate}
\begin{split}
    &\mathbf{H}^s_k=[\mathbf{h}^1_k;...;\mathbf{h}^\mathcal{T}_k], \\
    &\mathrm{Attn}(\mathbf{H}^s_k, \mathbf{e}_k) = \operatorname{Softmax}\left(\mathbf{H}^s_k\mathbf{e}_k^\top/{\sqrt{d}}\right), \\
    &\mathbf{s}_k 
     = \operatorname{Attn}(\mathbf{H}^s_k, \mathbf{e}_k)^{\top} \mathbf{H}^s_k,
\end{split}
\end{equation}
where $[;]$ denotes the row-wise concatenation.
Finally, we use an MLP layer to fuse the skill embedding and supply sequence representation to obtain the supply representation of the skill,
\begin{equation}
\label{eq:representation}
    \hat{\mathbf{s}}_k 
     = \operatorname{MLP}([\mathbf{e}_k || \mathbf{s}_k]\mathbf{W}_\mathcal{S}),
\end{equation}
where $\mathbf{W}_\mathcal{S}\in \mathbb{R}^{2d\times d}$ is a learnable parameter, $||$ denotes the column-wise concatenation.
Given the demand matrix $\mathbf{D}^{\mathcal{T}}_k$, we obtain skill demand representation $\hat{\mathbf{d}}_k$ in the same way.

\subsubsection{Cross-view relation learning.}

The cross-view relation learning block derives cross-view skill relationships between the supply and demand views by calculating the similarity score given the skill supply and demand representations.
The adaptive matrix  $\mathbf{A}_{p} \in \mathbb{R}^{2|\mathcal{K}| \times 2|\mathcal{K}|}$ is defined as

\begin{equation}
\begin{split}
&\mathbf{E}_\mathcal{S} = [\hat{\mathbf{s}}_1;\cdots;\hat{\mathbf{s}}_{|\mathcal{K}|}],~~ \mathbf{E}_\mathcal{D} = [\hat{\mathbf{d}}_1;\cdots;\hat{\mathbf{d}}_{|\mathcal{K}|}], \\
&\mathbf{X}_{A} = \tanh(\alpha [\mathbf{E}_\mathcal{S};\mathbf{E}_\mathcal{D}]),~~ \mathbf{X}_{B} = \tanh(\beta [\mathbf{E}_\mathcal{S};\mathbf{E}_\mathcal{D}]), \\
&\mathbf{A}_{p} = \sigma(\operatorname{Softmax}(\sigma(\mathbf{X}_{A}^\top\mathbf{X}_{B}-\mathbf{X}_{B}^\top\mathbf{X}_{A}))-\delta),
\end{split}
\label{eq:async adp}
\end{equation}
where $\mathbf{E}_\mathcal{S}$ and $\mathbf{E}_\mathcal{D}$ are the embedding matrices derived from the skill temporal encoding block, $\alpha, \beta \in \mathbb{R}$ denote learnable scalars, $\sigma$ stands for ReLU function, and $\delta$ is a hyperparameter to regulate the saturation of the adjacency matrix. 

\subsubsection{Cross-view augmentation.}

Leveraging the adaptive matrix $\mathbf{A}_{p}$, which encapsulates the cross-view asymmetric relationships, we further augment the cross-view skill demand and supply representations along with the intra-view skill relationships (\ie matrices $ \mathbf{A}^\mathcal{D}$ and $\mathbf{A}^\mathcal{S}$). We devise a two-layer graph convolution operation~\cite{kipf2016semi} to derive skill supply and demand representations. The cross-view representation is enhanced in line with the approach proposed by \cite{wu2019graph},

\begin{equation}
\label{eq:cross_equation}
\begin{gathered}
\mathbf{A}_{in} = 
\begin{bmatrix}
\mathbf{A}^\mathcal{S} & \mathbf{0} \\
\mathbf{0} & \mathbf{A}^\mathcal{D}
\end{bmatrix}, \\
[\tilde{\mathbf{E}}_\mathcal{S};\tilde{\mathbf{E}}_\mathcal{D}]  = \mathbf{A}_{p}[\mathbf{E}_\mathcal{S};\mathbf{E}_\mathcal{D}]\mathbf{W}_{p}  + \mathbf{A}_{in}[\mathbf{E}_\mathcal{S};\mathbf{E}_\mathcal{D}]\mathbf{W}_{in},
\end{gathered}
\end{equation}
where $[\tilde{\mathbf{E}}_\mathcal{S};\tilde{\mathbf{E}}_\mathcal{D}]$ denotes the learned embedding from cross-view graph encoder. 
$\mathbf{W}_{p},~\mathbf{W}_{in}\in \mathbb{R}^{d\times d}$ are learnable parameters. We denote $\tilde{\mathbf{E}}=[\tilde{\mathbf{E}}_\mathcal{S};\tilde{\mathbf{E}}_\mathcal{D}]$ in the following sections for the sake of simplicity.

\subsection{Hierarchical Graph Encoder (HGE)}

As aforementioned, the demand and supply of different skills may evolve synchronously. For example, the demand for \textit{C++} and \textit{Java} may boom under the policy of digital transformation. On the other hand, other skills, such as \textit{accounting}, may not share such a co-evolve pattern under the policy guidance. In this part, we introduce the hierarchical encoder to capture the high-level skill co-evolve patterns to further enhance the skill relationship modeling.

\subsubsection{Hierarchical clustering.}
To capture the cluster-wise trend among skills, we adopt Diffpool~\cite{ying2018hierarchical} to learn the cluster assignment in an end-to-end way.
Specifically, we define an soft assignment matrix, $\mathbf{S} \in \mathbb{R}^{2|\mathcal{K}|\times c}$, where
$c$ is a hyper-parameter determining the number of trend clusters. 
The assignment matrix is derived by
\begin{equation}
    \mathbf{S} = \operatorname{Softmax}(\tilde{\mathbf{E}}\mathbf{W}^\top_\mathbf{S}),
\end{equation}
where $\mathbf{S}$ describes the mapping relationship between bottom-layer skills and top-layer trend clusters,
$\mathbf{W}_\mathbf{S} \in \mathbb{R}^{c \times d}$ is denoted as a mapping function that maps the node to a specific cluster.
Then, we aggregate the connected skill representations to derive the cluster representations,
\begin{equation}
    \mathbf{X}_h = \mathbf{S}^\top \tilde{\mathbf{E}}.
\end{equation}

\subsubsection{Hierarchical augmentation.}

Then, we integrate the learned cluster representations back into the skill representations to enhance the skill representations.
\begin{equation}
\label{eq:hierarhical_equation}
\hat{\mathbf{E}}= \operatorname{Softmax}\left(\tilde{\mathbf{E}}\mathbf{X}_h^\top/\sqrt{d}\right) \mathbf{X}_h,
\end{equation}
where $\hat{\mathbf{E}} \in \mathbb{R}^{2|\mathcal{K}| \times d }$ denotes the learned skill representations enhanced by the hierarchical encoder, which including both $\hat{\mathbf{E}}_\mathcal{S}$ and $ \hat{\mathbf{E}}_\mathcal{D}$.

As clustering skills into appropriate trend clusters is a non-convex optimization problem, we introduce a clustering loss function~\cite{ying2018hierarchical},
\begin{equation}
\begin{gathered}
\mathcal{L}_{\text{cluster}} = \frac{1}{2|\mathcal{K}|}\sum_{i=1}^{2|\mathcal{K}|} \operatorname{H}(\mathbf{S}_i), \\
\operatorname{H}(\mathbf{S}_{i}) = -\sum_{j=0}^{c} \mathbf{S}_{i,j}\ln{\mathbf{S}}_{i,j},
\end{gathered}\label{loss:cluster}
\end{equation}
where $\operatorname{H}$ denotes the entropy functions and $\mathbf{S}_{i,j}$ denotes the probability that skill supply and demand trend $i$ belongs to the $j$-th trend cluster.
The loss function forces clear distinctions between intra-cluster and inter-cluster skill variations.

\subsection{Hyper-Decoder}
\newtheorem{example}{Example}

Despite the cross-view encoder and the hierarchical graph encoder preserving cross-view multi-level skill correlations, separately output skill demand and supply predictions may derive inaccurate skill gaps and lead to biased conclusions. We further elaborate on this problem with the following motivation example.

\begin{example}
Consider a machine learning engineer deciding on acquiring \textit{``Prompting''} skill in the forthcoming months.
If the predictive model underestimates the number of professionals who already have this skill, but correctly anticipates its rising demand, the engineer might face a saturated job market. 
Despite their investment in learning, they could struggle to find a competitive advantage for themselves to secure desired positions. 
\end{example}

In this work, we propose to incorporate historical skill gaps as auxiliary information to further calibrate supply and demand predictions.
Specifically, inspired by the success of hypernetwork in generating generalizable parameter weights based on specific conditions~\cite{pilault2020conditionally, Han2021joint}, we propose a conditional hyper-decoder to force the framework collectively make predictions by considering the demand-supply gap.
By encoding the prior skill gap sequence $ SG^\mathcal{T}_k$ as condition $\mathbf{c}^{t+1}_k$, we generate the weight $\Theta_k$ for the hyper-decoder as follows,
\begin{equation}
\begin{split}
&\mathbf{c}^{t+1}_k = \operatorname{LSTM}(SG^{t}_k,\mathbf{c}^{t}_k), \\
&\Theta_k = \operatorname{P}(\mathbf{c}^{t+1}_k, \Phi),
\end{split}
\end{equation}
where the LSTM layer transforms the skill gap sequence into conditions for supply and demand decoders, and $\mathbf{c}^0_k$ is initialized by skill embedding, $\mathbf{e}_k$. 
$\operatorname{P}(\cdot)$ is the hyper-network that generates the weight for the MLP decoder given the condition $\mathbf{c}^{t+1}_k$, and the gradients are computed with respect to the weight of hypernet $\Phi$. 
$\Theta_k$ is the generated decoder parameter for the skill $k$.

It's worth noting that updating the conditional hyper-decoder necessitates a carefully adjusted learning rate. 
An inappropriate learning rate may lead to slow convergence or exploding gradient issues.
Thus, we normalize the condition $\mathbf{c}^{t+1}_k$ and introduce another three-layer $\text{MLP}$ decoder as a supportive function to stabilize the training process,
\begin{equation}
\label{eq:hyper_equation}
\begin{gathered}
\hat{y}_k^\mathcal{S} = \operatorname{Softmax}(\operatorname{MLP}(\bar{\mathbf{e}}_k^\mathcal{S},\Theta_k) + \operatorname{MLP}(\bar{\mathbf{e}}_k^\mathcal{S})), \\ 
\hat{y}_k^\mathcal{D} = \operatorname{Softmax}(\operatorname{MLP}(\bar{\mathbf{e}}_k^\mathcal{D},\Theta_k) + \operatorname{MLP}(\bar{\mathbf{e}}_k^\mathcal{D})),
\end{gathered}
\end{equation}
where $\hat{y}_k^\mathcal{S},\hat{y}_k^\mathcal{D}$ represent our prediction goals for the next timestamp, and $\Theta_k$ is the replaced weight of $\text{MLP}$. 
$\bar{\mathbf{e}}_k^\mathcal{S}$ and $\bar{\mathbf{e}}_k^\mathcal{D}$ are the aggregated representations learned from cross-view graph encoder and hierarchical graph encoder, \ie $\bar{\mathbf{e}}_k^\mathcal{S} = \tilde{{\mathbf{e}}}_k^\mathcal{S} + \hat{\mathbf{e}}_k^\mathcal{S}$, $\bar{\mathbf{e}}_k^\mathcal{D} = \tilde{{\mathbf{e}}}_k^\mathcal{D} + \hat{\mathbf{e}}_k^\mathcal{D}$, 
where $\tilde{{\mathbf{e}}}_k^\mathcal{S}\in\tilde{{\mathbf{E}}}_\mathcal{S}$, $\tilde{{\mathbf{e}}}_k^\mathcal{D}\in \tilde{{\mathbf{E}}}_\mathcal{D}$, $\hat{\mathbf{e}}_k^\mathcal{S}\in \hat{\mathbf{E}}_\mathcal{S}$, $\hat{\mathbf{e}}_k^\mathcal{D}\in \hat{\mathbf{E}}_\mathcal{D} $.

\subsection{Optimization}

Following~\cite{guo2022talent}, we discretize the skill demand and supply into five trend categories, \ie \emph{high, medium-high, medium, medium-low, and low}. 
Please refer to \emph{Appendix} for a detailed processing procedure. 
The predictive loss is defined as
\begin{equation}
\mathcal{L}_{\text{main}} = -\frac{1}{2|\mathcal{K}|}\sum_{k=1}^{|\mathcal{K}|}\sum_{j=1}^{m}y^\mathcal{S}_{k,j} \log \hat{y}^\mathcal{S}_{k,j} + y^\mathcal{D}_{k,j}\log \hat{y}^\mathcal{D}_{k,j},
\end{equation}
where $y^\mathcal{D}_{k,j}$ and $y^\mathcal{S}_{k,j}$ are the ground truth labels of the skill $k$ and transformed from the ground truth supply and demand shares in time period $t+1$.
Take the demand label for example, $y^\mathcal{D}_{k}$ is a one-hot vector, and $y^\mathcal{D}_{k,j}=1$ if the demand share $\mathcal{D}_k^{t+1}$ locates at the $j$-th trend class, otherwise $y^\mathcal{D}_{k,j}=0$.
$m$ denotes the number of trend classes. 

The overall training objective of CHGH is to minimize 
\begin{equation}
\mathcal{L}  = \mathcal{L}_{\text{main}} + \lambda_1\mathcal{L}_{\text{cluster}} + \lambda_2||\theta||^2_2,
\end{equation}
where $\mathcal{L}_{\text{main}}$ is predictive loss, $\mathcal{L}_{\text{cluster}}$ is the cluster assignment loss as defined in Eq.~\eqref{loss:cluster}, and $||\theta||^2_2$ denotes the L2 regularization of learned parameters. $\lambda_1$ and $\lambda_2$ are the ratio of the clustering loss and the regularizer.

\section{Experiments}
In this section, we conduct extensive experiments to evaluate the effectiveness of our proposed framework. 
The source code of CHGH and all baselines are available online\footnote{https://github.com/vincent40416/Skill-Demand-Supply-Joint-Prediction}.
 
\subsection{Experimental Setup}

\begin{table*}[t]
\small
\centering
\begin{tabular}{l|cccc|cccc|cccc}
\toprule
Datasets & \multicolumn{4}{c|}{IT} & \multicolumn{4}{c|}{FIN} & \multicolumn{4}{c}{CONS} \\ 

\midrule 
Models & ACC & F1 & AUC &J-ACC & ACC & F1 & AUC&J-ACC & ACC & F1 & AUC&J-ACC \\
\midrule 	
ARIMA&	0.3294& 	0.3314& 	0.5809& 0.1161&	0.3226& 	0.3256& 	0.5766& 0.1072&	 0.3343& 	0.3375& 	0.5839& 0.1140 \\
VAR&	0.3324& 	0.3264& 	0.5827& 0.1170&	0.3190& 	0.3216& 	0.5744& 0.0997&	 0.3366& 	0.3424& 	0.5854& 0.1049 \\
LSTM&	0.4864& 	0.4812& 	0.7878&   0.2361&	0.5058& 	0.5014& 	0.7960&   0.2555&	0.4810& 	0.4751& 	0.7852& 0.2278 \\
Transformer & 0.4583& 	0.4544& 	0.7618&   0.1893&  0.4847& 	0.4818& 	0.7778&   0.2340&  0.3995& 	0.3927& 	0.7103& 0.1675 \\
Autoformer & 0.3899& 	0.3700& 	0.7125&  0.1540& 0.3593& 	0.3314& 	0.6940&   0.1312&  0.3834& 	0.3614& 	0.7104& 0.1476 \\
Wavenet& \underline{0.6170}& 	\underline{0.6163}& 	\underline{0.8867}&  0.3816&   0.5914& 	0.5919& 	0.8683&  0.3567&   \underline{0.6702}& 	\underline{0.6727}& 	\underline{0.9157}& \underline{0.4491}  \\
MTGNN&   0.6061& 	0.6040& 	0.8688&  \underline{0.3881}&   \underline{0.6621}& 	\underline{0.7044}& 	\underline{0.9209}&  \underline{0.4302}&   0.6086&     0.6085&     0.8658& 0.3522  \\
\textbf{Ours}&	\textbf{0.6777}& 	\textbf{0.6784}& 	\textbf{0.8968}& \textbf{0.4674} 
&\textbf{0.7227}& \textbf{0.7229} & \textbf{0.9234}& \textbf{0.5390}& \textbf{0.7704}& 	\textbf{0.7713}& 	\textbf{0.9398} &\textbf{0.6236}  \\				
\bottomrule

\end{tabular}
\caption{Overall performance comparison. The best results among all the models are highlighted in bold. The best baseline results are marked by underline.}
\label{overall performance}
\end{table*}

\begin{table}[t]
\small
\centering
\begin{tabular}{l|c|c|c|c}
\toprule 
Models & ACC& 	F1 &	AUC &	J-ACC \\
\midrule 
Static Graph&	0.5398& 	0.5378& 	0.8372& 	0.2944 \\
+ Adaptive Graph&	0.5724&	0.5684& 	0.8638& 	0.3333 \\
+ CGE&	0.6179& 	0.6160& 	0.8814& 	0.3980 \\
+ HGE& 0.6596& 	0.6563& 	\textbf{0.9082}& 	0.4447\\
+ Hyper-Decoder&	\textbf{0.6777}& 	\textbf{0.6784}& 	0.8968& 	\textbf{0.4674} \\
\bottomrule

\end{tabular}
\caption{Ablation study of each module on IT dataset.}
\label{joint prediction}
\end{table}
\subsubsection{Data Processing and Evaluation metrics.} 
For evaluation, we split each dataset described in Section~\ref{sec:data} into training, validation, and testing subsets with a ratio of 8:1:1. 
In our study, we formulate the demand and supply forecasting task as a classification problem. Our primary evaluation metrics include accuracy (ACC), weighted F1-score (F1), and the area under the receiver operating characteristic (AUC). 
We particularly underscore the importance of joint accuracy (J-ACC) as it represents the overall correctness of supply and demand predictions for a given skill. 
The definition of J-ACC is provided in \emph{Appendix}.
The hyper-parameters and implementation details are reported in \emph{Appendix}.

\subsubsection{Baselines.}
We compare CHGH with the following seven baselines and divide them into three classes: 
\begin{itemize}
\item Statistical Time Series Model: \textbf{ARIMA}~\cite{box1970distribution} is a classical time series prediction model by taking differencing and moving the average into auto-regression. \textbf{Vector Auto-Regressor}~\cite{stock2001vector} is a multivariate regressive model by capturing the relationship among variables.
\item Recurrent-based Model: 
\textbf{Long-Short Term Memory (LSTM)}~\cite{hochreiter1997long} introduces the forget gate and output gate to RNN to solve the vanishing gradient problem. 
\item Transformer-based model: \textbf{Transformer}~\cite{vaswani2017attention} adopts self-attention mechanism along with positional encoding.
\textbf{AutoFormer}~\cite{wu2021autoformer} introduces series decomposition blocks to capture series periodicity in long 
sequence prediction. 
\item Graph Neural Network based multivariate model: \textbf{ Wavenet}~\cite{wu2019graph} uses graph convolutions and temporal 1D convolutions for spatio-temporal relation modeling.
\textbf{MTGNN}~\cite{wu2020connecting}  exploits uni-directed dependencies among variables with mix-hop propagation to model the spatio-temporal dependencies. 
\end{itemize}

\subsection{Performance Comparison}
Table~\ref{overall performance} presents the overall performance comparison of various models across three different datasets: Information Technology (IT), Financial Industry (FIN), and Consumer Discretionary and Consumer Staples (CONS). The results lead us to the following observations.
First, our proposed model outperforms all the baseline models across all datasets on all the evaluation metrics, demonstrating the effectiveness of our CHGH framework on skill supply demand joint prediction. 
Moreover, traditional time series forecasting models yield lower performance across all datasets and metrics, and the models based on neural networks demonstrate a significant improvement over traditional models. Furthermore, it is worth noting that the Recurrent-based method outperforms the Transformer-based methods in these datasets since the Transformer-based methods mainly focus on tackling long-horizon auto-correlation while overlooking the trend of variables.
This outcome underscores the importance of the Recurrent-based model. 

Besides, the results of Wavenet and MTGNN outperform statistical- and recurrent-based models. This is attributed to their ability to propagate information through adaptive adjacency matrices grounded in skill embedding. Our proposed model further achieves noticeable improvement by preserving the demand-supply relationship and cluster-wise skill relationships, which were overlooked in previous studies.

\subsection{Ablation Study}
To delve deeper into the contributions of each component of our framework, we conducted an ablation study by gradually introducing designed modules into a basic model until forming the complete CHGH model.
The variants of the model are as follows:
\textbf{Static Graph} is the basic model which propagates the embedding learned from Skill Encoder based on  fixed co-occurrence matrices and uses the plain $\text{MLP}$ decoder.
\textbf{Adaptive Graph} introduces the adaptive graph learner to generate an adjacency matrix on demand and supply views separately.
\textbf{Cross-view Graph Encoder (CGE)} introduces asymmetric relations between demand and supply views to enhance skill representations (refer to Eq.~(\ref{eq:cross_equation})).
\textbf{Hierarchical Graph Encoder (HGE)} further involves cluster-wise skill trend information (refer to Eq.~(\ref{eq:hierarhical_equation})).
\textbf{Hyper-Decoder} introduces the Hyper-decoder conditioned on historical skill gap sequences to enhance the joint prediction (refer to Eq.~(\ref{eq:hyper_equation})). This represents the complete framework of CHGH.

As evident from Table~\ref{joint prediction}, while introducing static and adaptive graphs offers performance improvements over naive sequential models, it is the cross-view learning framework that significantly boosts prediction results. The cross-view learning framework, which interchanges the relations between the demand and supply views, enhances the accuracy of the adaptive graph.
The hierarchical graph encoder, designed to capture high-level trend information, further augments the accuracy.
Lastly, introducing the conditional hyper-decoder showcases a marked improvement, especially in joint prediction accuracy, which indicates the effectiveness of joint prediction conditioned on historical skill gaps. 
The ablation study on other datasets is in \emph{Appendix}.

\begin{figure}[t]
    \label{fig:parameter}
    \centering
    \subfloat[Effect of $\delta$ \label{fig:parameter:a}]{\includegraphics[width=0.22\textwidth]{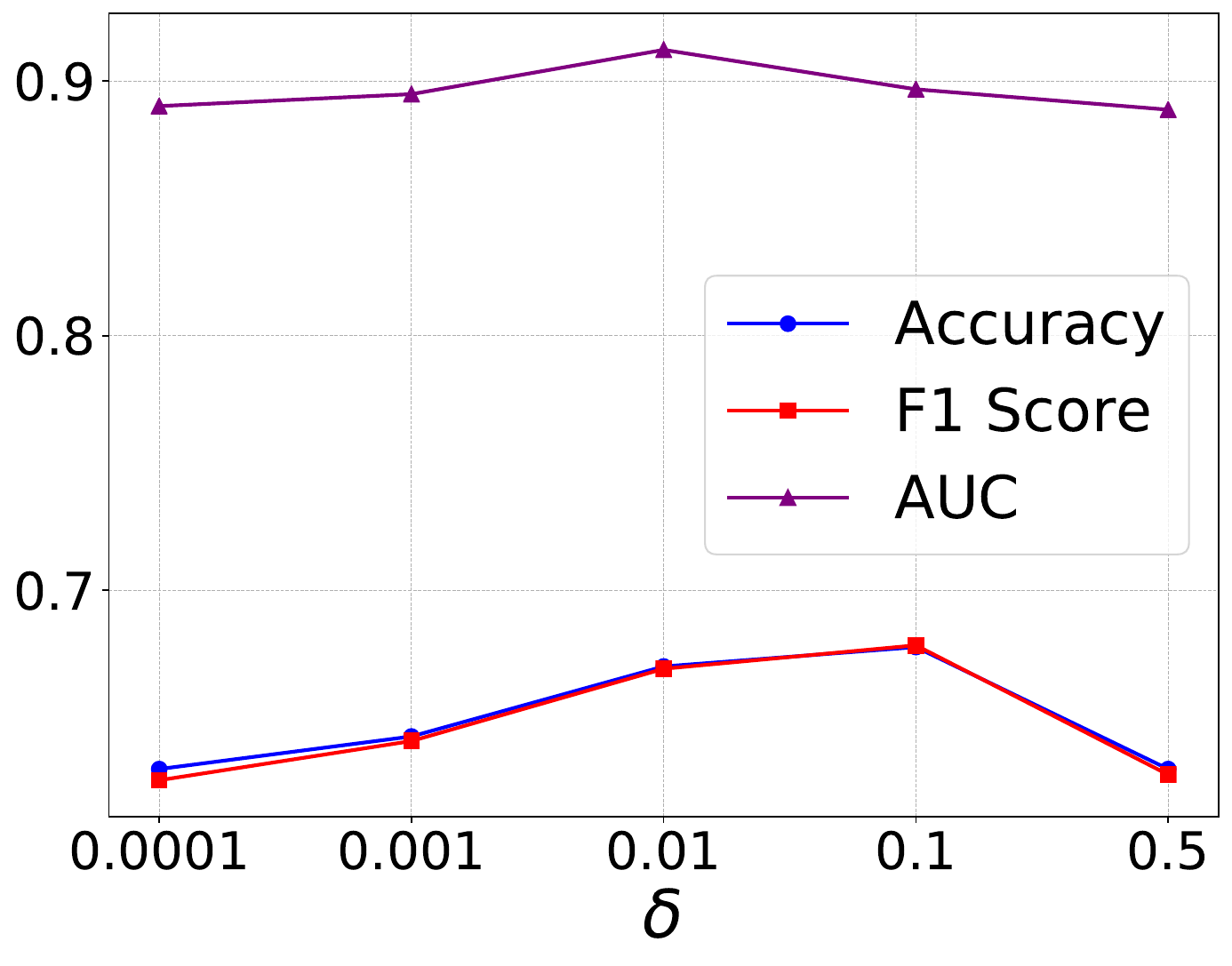}}
    \subfloat[Effect of $d$ \label{fig:parameter:b}]{\includegraphics[width=0.22\textwidth]{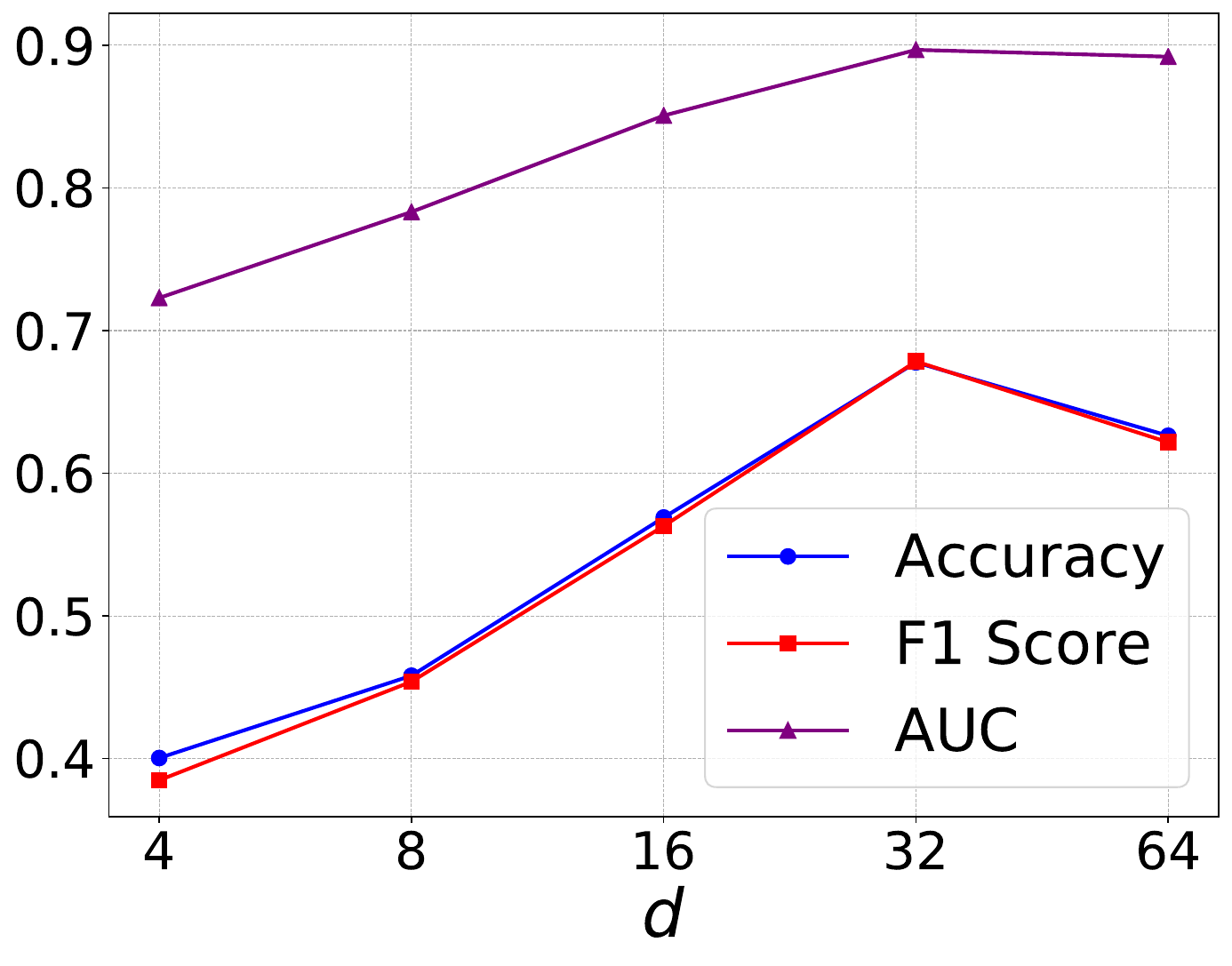}}
    \caption{Parameter sensitivity on IT dataset.}
\end{figure}

\subsection{Parameter Sensitivity}
\label{parameter sentivity}
We present a sensitivity analysis of CHGH. We report ACC, F1, and AUC on the IT dataset.

First, we vary $\delta$, which regulates the saturation of the edge generation, from $0.0001$ to $0.5$. A higher value implies that the generated edge becomes less. Figure~\ref{fig:parameter:a} illustrates that when setting too large $\delta$, the $\textbf{A}_{adp}$ is likely to become so sparse that there exists no connection between views. Besides, when setting $\delta$ too small, the connection will reach to near fully connected graph, also degrading the performance.

Then, we vary $d$, the dimension of the skill embedding, from $4$ to $64$. As shown in Figure~\ref{fig:parameter:b}, we observe that small dimension is insufficient to learn informative embedding, while high embedding dimension risks over-parameterization, causing the model to overfit to the dataset.

\subsection{Case Study}
\begin{figure}[t]
    \centering
    \includegraphics[width=0.85\linewidth]{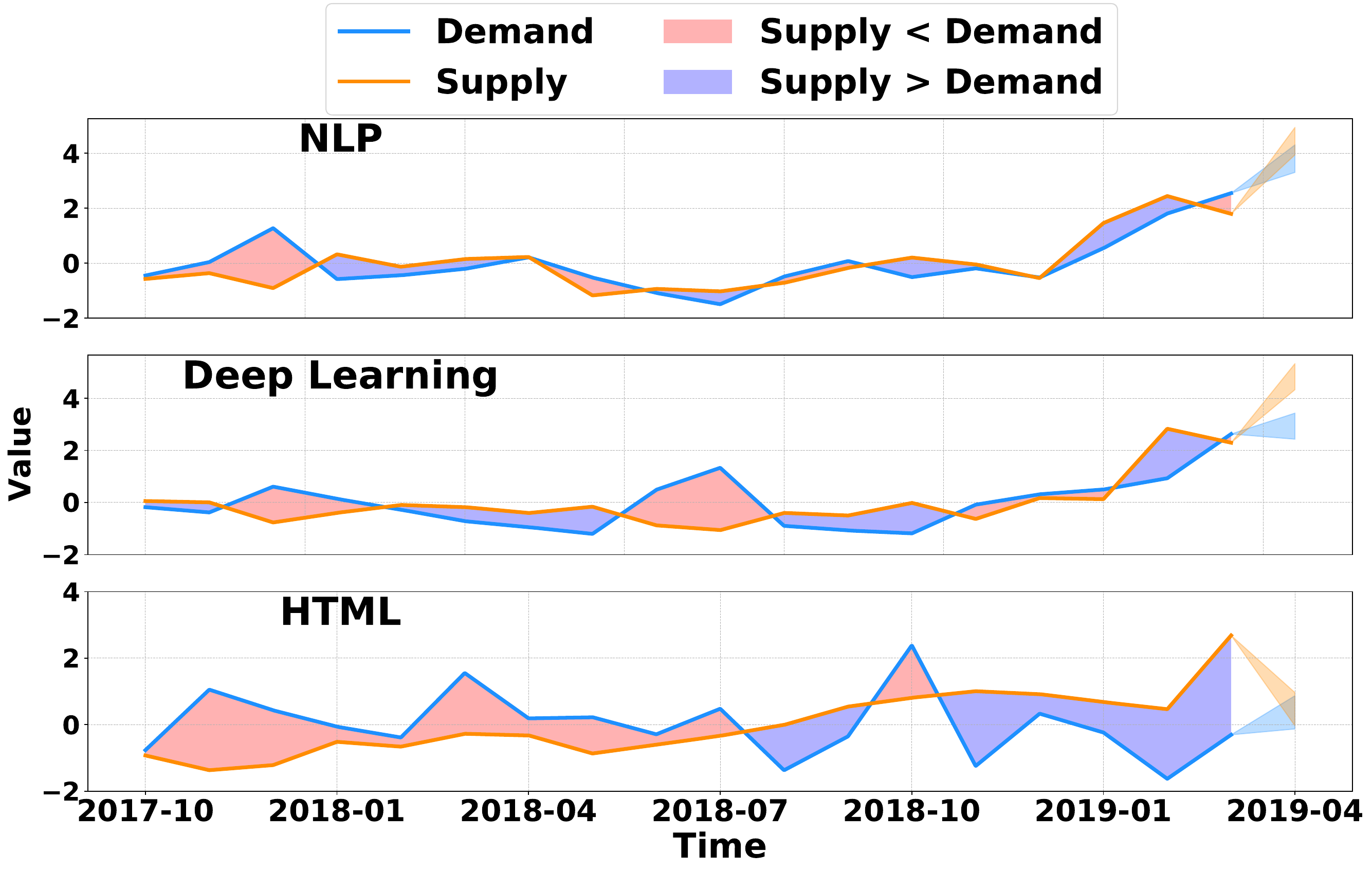}
    \caption{Visualization of ground truth and predicted skill demand and supply variations.}
    \label{fig:casestudy}
\end{figure}

As illustrated in Figure~\ref{fig:casestudy}, there is a noticeable increase in both supply and demand in the \textit{Natural Language Processing} (\textit{NLP}) domain around the time of BERT model publication by Google in October 2018~\cite{devlin2018bert}. 
Considering other concurrent factors that might have influenced the trend, many companies recognized the potential applications and advancements associated with BERT and other similar models, leading to increased interest in the field. 
Based on the prediction results of our model, there is an expected rise in both supply and demand for \textit{Deep Learning} and \textit{NLP} skills, which is consistent with observed real-world scenarios. 
Conversely, skills such as \textit{HTML}, primarily associated with web development, do not appear to follow the same trend as \textit{NLP} and \textit{Deep Learning}, facing an oversupply after October 2018. 
Overall, the study illustrates how our model facilitates users in selecting suitable skills to learn.

\section{Related Works}

\subsubsection{Skill supply and demand analysis.}
Traditional skill demand-supply analysis primarily involves statistical methods with expert knowledge to understand the pattern of transition of skills.
Some researchers analyzed the skill demand shift caused by technology advancement~\cite{bughin2018skill} and identified skill obsolescence~\cite{de2002economics} based on surveys. 
Due to the increasing usable data resources from online recruitment services, machine-learning-based models emerged as a powerful tool for substituting traditional statistic models. \citet{xu2018measuring} measures the popularity of skills with the topic model across various job criteria, such as company size and salary.  \citet{garcia2022practical} adopted recurrent neural network in skill demand prediction.
Despite the existence of research on skill-level supply~\cite{belhaj2013markov} and demand prediction~\cite{garcia2022practical}, they fail to address the connection between demand-supply in the predicting model.
Besides, there are studies that delve into company-position-level analyses of supply and demand. These research efforts model the competitive nature of recruitment strategies employed by various companies~\cite{wu2019trend,zhang2021talent, guo2022talent, qin2023comprehensive}.

\subsubsection{Graph neural network.}
Graph neural networks received significant attention from academia and industry owing to their capability on learning non-Euclidean relationships~\cite{shih2019temporal, wu2023learning, zheng2021drug, zheng2023interaction}. The existing work could be divided into two categories, spectral-based methods~\cite{defferrard2016convolutional,wu2019graph} and spatial-based methods~\cite{kipf2016semi,hamilton2017inductive,velickovic2017graph}. The former adopted the message-filtering process and the latter follow the message-passing rule to aggregate information. 
In this work, we employ the GNN to facilitate information propagation across views and capture skill variation correlations.

\section{Conclusion}
In this work, we investigate the skill demand-supply joint prediction task, empowering employees and employers to anticipate future skill demand and supply variations and address potential skill gaps accordingly.
To tackle the inherent challenges of this task, we proposed the CHGH framework. 
Specifically, a cross-view graph encoder is proposed to capture the inherent relationships between the skill supply and demand graph views.
Furthermore, a hierarchical module was used to identify and preserve high-level skill trends that were caused by technological advancements or policy changes.
We also constructed a conditioned hyper-decoder to calibrate supply and demand predictions by exploiting historical skill gaps as auxiliary signals. 
Extensive experiments demonstrated the superiority of CHGH against seven baselines, and the ablation study showed the effectiveness of each proposed module.
By bridging the gap between employees' skills and industry requirements, this work facilitates a better alignment between individuals and the evolving labor market requirements.

\appendix
\begin{appendices}
\section{Mathematical Notation}

{
\small
\begin{tabular}{l|p{0.36\textwidth}}
\toprule 
\textbf{Symbol} & \textbf{Description} \\
\midrule
$\mathcal{K}$ & Skill sets. \\
$t$ & Timestamp. \\
$\mathcal{T}$ & Time range $T = \{0,1,...t\}$. \\
$\mathcal{J}_t$ & Job description at time $t$. \\
$\mathcal{W}_t$ & Work experience at time $t$. \\
$\mathcal{D}^t_k$ & Demand share of skill $k$ at time $t$. \\
$\mathcal{S}^t_k$ & Supply share of skill $k$ at time $t$.  \\
$\mathcal{SG}^t_k$ & Skill gap of skill $k$ at time $t$.  \\
$\mathcal{G}_{\mathcal{D}}$ & Skill demand relation graph. \\
$\mathbf{A}^{\mathcal{D}}$ & Adjacency derived from $\mathcal{G}_{\mathcal{D}}$. \\
$\mathcal{G}_{\mathcal{S}}$ & Skill supply relation graph. \\
$\mathbf{A}^{\mathcal{S}}$ & Adjacency matrix derived from $\mathcal{G}_{\mathcal{S}}$. \\
$\hat{Y}^{t+1}_{\mathcal{D}}$ & Estimated target of demand share \\
$\hat{Y}^{t+1}_{\mathcal{S}}$ & Estimated target of supply share \\
$\mathbf{W}_*$ & Model parameters \\
$\mathbf{E}$ & Embedding derived from skill encoder. \\
$\mathbf{A}_{p}$ & Learned adjacency matrix\\
$\mathbf{A}_{in}$ & Combined intra-view adjacency matrix \\
$\tilde{\mathbf{E}}$ & Skill embedding derived from cross-view graph encoder. \\
$c$ & Number of clusters. \\
$\mathbf{S}$ & Assignment matrix. \\
$\mathbf{X}_h$ & Trend embedding. \\
$\hat{\mathbf{E}}$ &  Skill embedding derived from hierarchical graph encoder. \\
$\mathbf{c}^{t+1}_k$ &  Conditions of hyper-decoder. \\
$\bar{\mathbf{e}}_k^{\mathcal{D}}$ &Demand embedding as input of hyper-decoder. \\
$\bar{\mathbf{e}}_k^{\mathcal{S}}$ &Supply embedding as input of hyper-decoder. \\
$y_k^\mathcal{D}$ & Predicted demand share of skill $k$. \\
$y_k^\mathcal{S}$ & Predicted supply share of skill $k$. \\
\bottomrule

\end{tabular} 
}

\section{Discretizing Algorithm} \label{appendix:sax}
The algorithm maps the demand share prediction goal $y^{\mathcal{D}}_k$ to discrete classes, the detail of the algorithm is formulated as: 
\begin{equation}
\begin{gathered}
std_k = STD(\mathcal{D}^{\mathcal{T}}_k), \\
\bar{\mathcal{D}}_k = \sum_{i=1}^{t}{\mathcal{D}}^i_k, \\
\hat{\mathcal{D}}^{t+1}_k = \frac{\mathcal{D}^{t+1}_k-\bar{\mathcal{D}}_k}{std_k}, \\
\hat{\mathcal{D}}^{t+1}_\mathcal{K} = \{\hat{\mathcal{D}}^{t+1}_0,\hat{\mathcal{D}}^{t+1}_1, \dots ,\hat{\mathcal{D}}^{t+1}_{|\mathcal{K}|}\}, \\
\end{gathered}
\end{equation}
where $STD(*)$ calculates the standard deviation of the sequence.
After sorting $\hat{\mathcal{D}}^{t+1}_\mathcal{K}$ in ascending order, as  $\tilde{\mathcal{D}}^{t+1}_\mathcal{K}$, we divide them into \( n \) classes with an equal number of elements in each class. The classes is represented as \( C_1, C_2, \dots, C_n \). which is formulated as: 
\begin{equation}
\begin{gathered}
\tilde{\mathcal{D}}^{t+1}_\mathcal{K} = \{\tilde{\mathcal{D}}^{t+1}_0, \tilde{\mathcal{D}}^{t+1}_1, \dots, \tilde{\mathcal{D}}^{t+1}_{|\mathcal{K}|}\}, \\
\tilde{\mathcal{D}}^{t+1}_0 \leq \tilde{\mathcal{D}}^{t+1}_1 \leq \dots \leq \tilde{\mathcal{D}}^{t+1}_{|\mathcal{K}|}, \\
C_i = \{\tilde{\mathcal{D}}^{t+1}_{(i-1) \times \frac{n}{k}}, \tilde{\mathcal{D}}^{t+1}_{(i-1) \times \frac{n}{k} + 1}, \dots, \tilde{\mathcal{D}}^{t+1}_{i \times \frac{n}{k} - 1}\},
\end{gathered}
\end{equation}
where $n$ is set to five trends in our experiment, representing \emph{high, medium-high, medium, medium-low, and low}.
The ground truth $y^\mathcal{D}_{k,i}=1$ if $\hat{\mathcal{D}}^{t+1}_k \in C_i$, otherwise $y^\mathcal{D}_{k,i}=0 $.
$y^{\mathcal{S}}_k$ is discretized in a similar way based on the supply share sequence.

\section{Joint Accuracy}
\label{appendix:joint accuracy}
To verify that both supply and demand are correctly predicted. 
We use \( y^{\mathcal{S}}_{k} \) and \( \hat{y}^{\mathcal{S}}_{k} \) as the true and predicted labels for the supply of the \( k^{th} \) skill, \( y^{\mathcal{D}}_{k} \) and \( \hat{y}^{\mathcal{D}}_{k} \) be the true and predicted labels for the \( D \) goal of the \( k^{th} \) instance.\( |\mathcal{K}| \) is the total number of instances.
Define the indicator function \( I \) as:
\begin{equation}
I(y^{\mathcal{S}}_{k}, \hat{y}^{\mathcal{S}}_{k}, y^{\mathcal{D}}_{k}, \hat{y}^{\mathcal{D}}_{k}) = 
\begin{cases} 
    1 & \text{if } y^{\mathcal{S}}_{k} = \hat{y}^{\mathcal{S}}_{k} \text{ and } y^{\mathcal{D}}_{k} =  \hat{y}^{\mathcal{D}}_{k} \\
    0 & \text{otherwise}
\end{cases}.
\end{equation}
The joint accuracy (J-ACC) is then given by:
\begin{equation}
ACC = \frac{\sum_{k=1}^{|\mathcal{K}|} I(y^{\mathcal{S}}_{k}, \hat{y}^{\mathcal{S}}_{k}, y^{\mathcal{D}}_{k}, \hat{y}^{\mathcal{D}}_{k})}{|\mathcal{K}|}.
\end{equation}

\section{Implementation Details} \label{appendix:Experimental setup}

For hyper-parameters, we choose the number of trend types as 5, the minimum length of sequences as 5, the embedding dim $d$ as 32, the number of clusters $c$ is 100, the head number of multi-head attention as 4, the number of $\text{LSTM}$ layers as 3, and the output dimension of any other multi-layer perceptron as 32. 
We use Adam Optimization with a learning rate of 1e-3.
The dropout is set to 0.3. 
The reducing rate of the learning rate scheduler is 0.1, step as 50. 
$\lambda_1, \lambda_2$ are set as 1e-5.
We run each method 5 times and report the average results. 
The model is run on the machine with Intel(R) Xeon(R) Gold 5118 CPU @ 2.30GHz, NVIDIA GeForce RTX 3090 with 24G memory. The operating system is Centos Linux 7 (Core).
\begin{table}[t]
\small
\centering
\resizebox{\linewidth}{!}{
\begin{tabular}{l|c|c|c|c}
\toprule 
Models & ACC& 	F1 &	AUC &	J-ACC \\
\midrule

Static Graph& 0.4546 	&0.4348 	&0.7771 	&0.2230 \\ 
+ Adaptive Graph& 0.5826 	&0.5784 	&0.8627 	&0.3615 \\ 
+ CGE & 0.6839 	&0.6831 	&0.9135 	&0.4833 \\ 
+ HGE & 0.7218 	&0.7220 	&0.9170 	&0.5158 \\
+ Hyper-Decoder& \textbf{0.7227} 	&\textbf{0.7229} 	&\textbf{0.9234} 	&\textbf{0.5390} \\

\bottomrule
\end{tabular}

}
\caption{Ablation study of each module on FIN dataset.}
\label{appendix:ablation study on Fin}
\end{table}

\begin{table}[t]
\small
\centering
\resizebox{\linewidth}{!}{
\begin{tabular}{l|c|c|c|c}
\toprule 
Models & ACC& 	F1 &	AUC &	J-ACC \\
\midrule 
Static Graph& 0.5183 	&0.5135 	&0.8177 	&0.2773 \\
+ Adaptive Graph& 0.5660 	&0.5628 	&0.8509 	&0.3242 \\ 
+ CGE & 0.7167 	&0.7172 	&0.9237 	&0.5414 \\ 
+ HGE & 0.7440 	&0.7441 	&0.9328 	&0.5800 \\
+ Hyper-Decoder& \textbf{0.7704} 	&\textbf{0.7713} 	&\textbf{0.9398} 	&\textbf{0.6236} \\
\bottomrule
\end{tabular}

}
\caption{Ablation study of each module on CONS dataset.}
\label{appendix:ablation study on Cons}
\end{table}

\section{Ablation Study}
\label{appendix:Ablation Study}
The ablation study on the other dataset is shown in Table~\ref{appendix:ablation study on Fin} and~\ref{appendix:ablation study on Cons}. The results show the same tendency as discussed in the study, indicating the effectiveness of the designed module across all the datasets.

\end{appendices}

\section*{Acknowledgements}

This research was partly supported by the National Natural Science Foundation of China under Grant No. 62102110, 92370204, Guangzhou Science and Technology Plan Guangzhou-HKUST(GZ) Joint Project No. 2023A03J0144.
And this work was partially done at Boss Zhipin Career Science Lab.

\bibliography{aaai24}

\begin{thebibliography}{38}
\providecommand{\natexlab}[1]{#1}

\bibitem[{Belhaj and Tkiouat(2013)}]{belhaj2013markov}
Belhaj, R.; and Tkiouat, M. 2013.
\newblock A Markov Model for Human Resources Supply Forecast Dividing the HR System into Subgroups.
\newblock \emph{Journal of Service Science and Management}, 6: 211--217.

\bibitem[{Box and Pierce(1970)}]{box1970distribution}
Box, G.~E.; and Pierce, D.~A. 1970.
\newblock Distribution of residual autocorrelations in autoregressive-integrated moving average time series models.
\newblock \emph{Journal of the American statistical Association}, 65(332): 1509--1526.

\bibitem[{Bughin et~al.(2018)Bughin, Hazan, Lund, Dahlstr{\"o}m, Wiesinger, and Subramaniam}]{bughin2018skill}
Bughin, J.; Hazan, E.; Lund, S.; Dahlstr{\"o}m, P.; Wiesinger, A.; and Subramaniam, A. 2018.
\newblock Skill shift: Automation and the future of the workforce.
\newblock \emph{McKinsey Global Institute}, 1: 3--84.

\bibitem[{Communications(2022)}]{LinkedInSkill}
Communications, L.~C. 2022.
\newblock Our skills-first vision for the future.
\newblock \url{https://economicgraph.linkedin.com/blog/a-skills-first-blueprint-for-better-job-outcomes}.
\newblock Accessed: 2022-03-29.

\bibitem[{De~Grip and Van~Loo(2002)}]{de2002economics}
De~Grip, A.; and Van~Loo, J. 2002.
\newblock The economics of skills obsolescence: a review.
\newblock \emph{Research in Labor Economics}, 21: 1--26.

\bibitem[{de~Macedo et~al.(2022)de~Macedo, Clarke, Lucherini, Baldwin, Neto, de~Paula, and Das}]{garcia2022practical}
de~Macedo, M. M.~G.; Clarke, W.; Lucherini, E.; Baldwin, T.; Neto, D.~Q.; de~Paula, R.~A.; and Das, S. 2022.
\newblock Practical Skills Demand Forecasting via Representation Learning of Temporal Dynamics.
\newblock In Conitzer, V.; Tasioulas, J.; Scheutz, M.; Calo, R.; Mara, M.; and Zimmermann, A., eds., \emph{{AIES} '22: {AAAI/ACM} Conference on AI, Ethics, and Society, Oxford, United Kingdom, May 19 - 21, 2021}, 285--294. {ACM}.

\bibitem[{Defferrard, Bresson, and Vandergheynst(2016)}]{defferrard2016convolutional}
Defferrard, M.; Bresson, X.; and Vandergheynst, P. 2016.
\newblock Convolutional Neural Networks on Graphs with Fast Localized Spectral Filtering.
\newblock In Lee, D.~D.; Sugiyama, M.; von Luxburg, U.; Guyon, I.; and Garnett, R., eds., \emph{Advances in Neural Information Processing Systems 29: Annual Conference on Neural Information Processing Systems 2016, December 5-10, 2016, Barcelona, Spain}, 3837--3845.

\bibitem[{Devlin et~al.(2018)Devlin, Chang, Lee, and Toutanova}]{devlin2018bert}
Devlin, J.; Chang, M.; Lee, K.; and Toutanova, K. 2018.
\newblock {BERT:} Pre-training of Deep Bidirectional Transformers for Language Understanding.
\newblock \emph{CoRR}, abs/1810.04805.

\bibitem[{Donovan et~al.(2022)Donovan, Stoll, Bradley, and Collins}]{donovan2022skills}
Donovan, S.~A.; Stoll, A.; Bradley, D.~H.; and Collins, B. 2022.
\newblock Skills Gaps: A Review of Underlying Concepts and Evidence. CRS Report R47059, Version 3.
\newblock \emph{Congressional Research Service}.

\bibitem[{Guo et~al.(2022)Guo, Liu, Zhang, Zhang, Zhu, and Xiong}]{guo2022talent}
Guo, Z.; Liu, H.; Zhang, L.; Zhang, Q.; Zhu, H.; and Xiong, H. 2022.
\newblock Talent Demand-Supply Joint Prediction with Dynamic Heterogeneous Graph Enhanced Meta-Learning.
\newblock In Zhang, A.; and Rangwala, H., eds., \emph{{KDD} '22: The 28th {ACM} {SIGKDD} Conference on Knowledge Discovery and Data Mining, Washington, DC, USA, August 14 - 18, 2022}, 2957--2967. {ACM}.

\bibitem[{Hamilton, Ying, and Leskovec(2017)}]{hamilton2017inductive}
Hamilton, W.~L.; Ying, Z.; and Leskovec, J. 2017.
\newblock Inductive Representation Learning on Large Graphs.
\newblock In Guyon, I.; von Luxburg, U.; Bengio, S.; Wallach, H.~M.; Fergus, R.; Vishwanathan, S. V.~N.; and Garnett, R., eds., \emph{Advances in Neural Information Processing Systems 30: Annual Conference on Neural Information Processing Systems 2017, December 4-9, 2017, Long Beach, CA, {USA}}, 1024--1034.

\bibitem[{Han et~al.(2021)Han, Liu, Zhu, Xiong, and Dou}]{Han2021joint}
Han, J.; Liu, H.; Zhu, H.; Xiong, H.; and Dou, D. 2021.
\newblock Joint Air Quality and Weather Prediction Based on Multi-Adversarial Spatiotemporal Networks.
\newblock In \emph{Thirty-Fifth {AAAI} Conference on Artificial Intelligence, {AAAI} 2021, Thirty-Third Conference on Innovative Applications of Artificial Intelligence, {IAAI} 2021, The Eleventh Symposium on Educational Advances in Artificial Intelligence, {EAAI} 2021, Virtual Event, February 2-9, 2021}, 4081--4089. {AAAI} Press.

\bibitem[{Hochreiter and Schmidhuber(1997)}]{hochreiter1997long}
Hochreiter, S.; and Schmidhuber, J. 1997.
\newblock Long short-term memory.
\newblock \emph{Neural computation}, 9(8): 1735--1780.

\bibitem[{Khan and Blumenstock(2019)}]{khan2019multi}
Khan, M.~R.; and Blumenstock, J.~E. 2019.
\newblock Multi-GCN: Graph Convolutional Networks for Multi-View Networks, with Applications to Global Poverty.
\newblock In \emph{The Thirty-Third {AAAI} Conference on Artificial Intelligence, {AAAI} 2019, The Thirty-First Innovative Applications of Artificial Intelligence Conference, {IAAI} 2019, The Ninth {AAAI} Symposium on Educational Advances in Artificial Intelligence, {EAAI} 2019, Honolulu, Hawaii, USA, January 27 - February 1, 2019}, 606--613. {AAAI} Press.

\bibitem[{Kim, Hsu, and Stern(2006)}]{kim2006update}
Kim, Y.; Hsu, J.; and Stern, M. 2006.
\newblock An update on the IS/IT skills gap.
\newblock \emph{Journal of information systems education}, 17(4): 395.

\bibitem[{Kipf and Welling(2016)}]{kipf2016semi}
Kipf, T.~N.; and Welling, M. 2016.
\newblock Semi-Supervised Classification with Graph Convolutional Networks.
\newblock \emph{CoRR}, abs/1609.02907.

\bibitem[{Larsen et~al.(2018)Larsen, Rand, Schmid, and Dean}]{larsen2018developing}
Larsen, C.; Rand, S.; Schmid, A.; and Dean, A. 2018.
\newblock \emph{Developing Skills in a Changing World of Work: concepts, measurement and data applied in regional and local labour market monitoring across Europe}.
\newblock Rainer Hampp Verlag Augsburg/M{\"u}nchen, Germany.

\bibitem[{Liang et~al.(2020)Liang, Huang, Wang, and Yu}]{liang2022multi}
Liang, Y.; Huang, D.; Wang, C.; and Yu, P.~S. 2020.
\newblock Multi-view Graph Learning by Joint Modeling of Consistency and Inconsistency.
\newblock \emph{CoRR}, abs/2008.10208.

\bibitem[{McGuinness and Ortiz(2016)}]{mcguinness2016skill}
McGuinness, S.; and Ortiz, L. 2016.
\newblock Skill gaps in the workplace: measurement, determinants and impacts.
\newblock \emph{Industrial relations journal}, 47(3): 253--278.

\bibitem[{Phillips and Ormsby(2016)}]{phillips2016industry}
Phillips, R.~L.; and Ormsby, R. 2016.
\newblock Industry classification schemes: An analysis and review.
\newblock \emph{Journal of Business and Finance Librarianship}, 21(1): 1--25.

\bibitem[{Pilault, Elhattami, and Pal(2021)}]{pilault2020conditionally}
Pilault, J.; Elhattami, A.; and Pal, C.~J. 2021.
\newblock Conditionally Adaptive Multi-Task Learning: Improving Transfer Learning in {NLP} Using Fewer Parameters {\&} Less Data.
\newblock In \emph{9th International Conference on Learning Representations, {ICLR} 2021, Virtual Event, Austria, May 3-7, 2021}. OpenReview.net.

\bibitem[{Qin et~al.(2023)Qin, Zhang, Zha, Shen, Zhang, Sun, Zhu, Zhu, and Xiong}]{qin2023comprehensive}
Qin, C.; Zhang, L.; Zha, R.; Shen, D.; Zhang, Q.; Sun, Y.; Zhu, C.; Zhu, H.; and Xiong, H. 2023.
\newblock A Comprehensive Survey of Artificial Intelligence Techniques for Talent Analytics.
\newblock \emph{CoRR}, abs/2307.03195.

\bibitem[{Radford et~al.(2018)Radford, Narasimhan, Salimans, Sutskever et~al.}]{radford2018improving}
Radford, A.; Narasimhan, K.; Salimans, T.; Sutskever, I.; et~al. 2018.
\newblock Improving language understanding by generative pre-training.

\bibitem[{Shih, Sun, and Lee(2019)}]{shih2019temporal}
Shih, S.; Sun, F.; and Lee, H. 2019.
\newblock Temporal pattern attention for multivariate time series forecasting.
\newblock \emph{Machine Learning}, 108(8-9): 1421--1441.

\bibitem[{Stock and Watson(2001)}]{stock2001vector}
Stock, J.~H.; and Watson, M.~W. 2001.
\newblock Vector autoregressions.
\newblock \emph{Journal of Economic perspectives}, 15(4): 101--115.

\bibitem[{Vaswani et~al.(2017)Vaswani, Shazeer, Parmar, Uszkoreit, Jones, Gomez, Kaiser, and Polosukhin}]{vaswani2017attention}
Vaswani, A.; Shazeer, N.; Parmar, N.; Uszkoreit, J.; Jones, L.; Gomez, A.~N.; Kaiser, L.; and Polosukhin, I. 2017.
\newblock Attention is All you Need.
\newblock In Guyon, I.; von Luxburg, U.; Bengio, S.; Wallach, H.~M.; Fergus, R.; Vishwanathan, S. V.~N.; and Garnett, R., eds., \emph{Advances in Neural Information Processing Systems 30: Annual Conference on Neural Information Processing Systems 2017, December 4-9, 2017, Long Beach, CA, {USA}}, 5998--6008.

\bibitem[{Velickovic et~al.(2017)Velickovic, Cucurull, Casanova, Romero, Li{\`{o}}, and Bengio}]{velickovic2017graph}
Velickovic, P.; Cucurull, G.; Casanova, A.; Romero, A.; Li{\`{o}}, P.; and Bengio, Y. 2017.
\newblock Graph Attention Networks.
\newblock \emph{CoRR}, abs/1710.10903.

\bibitem[{Wang et~al.(2016)Wang, Huang, Zhu, and Zhao}]{wang2016attention}
Wang, Y.; Huang, M.; Zhu, X.; and Zhao, L. 2016.
\newblock Attention-based {LSTM} for Aspect-level Sentiment Classification.
\newblock In Su, J.; Carreras, X.; and Duh, K., eds., \emph{Proceedings of the 2016 Conference on Empirical Methods in Natural Language Processing, {EMNLP} 2016, Austin, Texas, USA, November 1-4, 2016}, 606--615. The Association for Computational Linguistics.

\bibitem[{Wu et~al.(2021)Wu, Xu, Wang, and Long}]{wu2021autoformer}
Wu, H.; Xu, J.; Wang, J.; and Long, M. 2021.
\newblock Autoformer: Decomposition Transformers with Auto-Correlation for Long-Term Series Forecasting.
\newblock In Ranzato, M.; Beygelzimer, A.; Dauphin, Y.~N.; Liang, P.; and Vaughan, J.~W., eds., \emph{Advances in Neural Information Processing Systems 34: Annual Conference on Neural Information Processing Systems 2021, NeurIPS 2021, December 6-14, 2021, virtual}, 22419--22430.

\bibitem[{Wu et~al.(2023)Wu, Zhao, Li, Huang, Liu, and Chen}]{wu2023learning}
Wu, L.; Zhao, H.; Li, Z.; Huang, Z.; Liu, Q.; and Chen, E. 2023.
\newblock Learning the Explainable Semantic Relations via Unified Graph Topic-Disentangled Neural Networks.
\newblock \emph{ACM Transactions on Knowledge Discovery from Data}, 17(8): 1--23.

\bibitem[{Wu et~al.(2019{\natexlab{a}})Wu, Xu, Zhu, Zhang, Chen, and Xiong}]{wu2019trend}
Wu, X.; Xu, T.; Zhu, H.; Zhang, L.; Chen, E.; and Xiong, H. 2019{\natexlab{a}}.
\newblock Trend-Aware Tensor Factorization for Job Skill Demand Analysis.
\newblock In Kraus, S., ed., \emph{Proceedings of the Twenty-Eighth International Joint Conference on Artificial Intelligence, {IJCAI} 2019, Macao, China, August 10-16, 2019}, 3891--3897. ijcai.org.

\bibitem[{Wu et~al.(2020)Wu, Pan, Long, Jiang, Chang, and Zhang}]{wu2020connecting}
Wu, Z.; Pan, S.; Long, G.; Jiang, J.; Chang, X.; and Zhang, C. 2020.
\newblock Connecting the Dots: Multivariate Time Series Forecasting with Graph Neural Networks.
\newblock In Gupta, R.; Liu, Y.; Tang, J.; and Prakash, B.~A., eds., \emph{{KDD} '20: The 26th {ACM} {SIGKDD} Conference on Knowledge Discovery and Data Mining, Virtual Event, CA, USA, August 23-27, 2020}, 753--763. {ACM}.

\bibitem[{Wu et~al.(2019{\natexlab{b}})Wu, Pan, Long, Jiang, and Zhang}]{wu2019graph}
Wu, Z.; Pan, S.; Long, G.; Jiang, J.; and Zhang, C. 2019{\natexlab{b}}.
\newblock Graph WaveNet for Deep Spatial-Temporal Graph Modeling.
\newblock In Kraus, S., ed., \emph{Proceedings of the Twenty-Eighth International Joint Conference on Artificial Intelligence, {IJCAI} 2019, Macao, China, August 10-16, 2019}, 1907--1913. ijcai.org.

\bibitem[{Xu et~al.(2018)Xu, Zhu, Zhu, Li, and Xiong}]{xu2018measuring}
Xu, T.; Zhu, H.; Zhu, C.; Li, P.; and Xiong, H. 2018.
\newblock Measuring the Popularity of Job Skills in Recruitment Market: {A} Multi-Criteria Approach.
\newblock In McIlraith, S.~A.; and Weinberger, K.~Q., eds., \emph{Proceedings of the Thirty-Second {AAAI} Conference on Artificial Intelligence, (AAAI-18), the 30th innovative Applications of Artificial Intelligence (IAAI-18), and the 8th {AAAI} Symposium on Educational Advances in Artificial Intelligence (EAAI-18), New Orleans, Louisiana, USA, February 2-7, 2018}, 2572--2579. {AAAI} Press.

\bibitem[{Ying et~al.(2018)Ying, You, Morris, Ren, Hamilton, and Leskovec}]{ying2018hierarchical}
Ying, Z.; You, J.; Morris, C.; Ren, X.; Hamilton, W.~L.; and Leskovec, J. 2018.
\newblock Hierarchical Graph Representation Learning with Differentiable Pooling.
\newblock In Bengio, S.; Wallach, H.~M.; Larochelle, H.; Grauman, K.; Cesa{-}Bianchi, N.; and Garnett, R., eds., \emph{Advances in Neural Information Processing Systems 31: Annual Conference on Neural Information Processing Systems 2018, NeurIPS 2018, December 3-8, 2018, Montr{\'{e}}al, Canada}, 4805--4815.

\bibitem[{Zhang et~al.(2021)Zhang, Zhu, Sun, Liu, Zhuang, and Xiong}]{zhang2021talent}
Zhang, Q.; Zhu, H.; Sun, Y.; Liu, H.; Zhuang, F.; and Xiong, H. 2021.
\newblock Talent Demand Forecasting with Attentive Neural Sequential Model.
\newblock In Zhu, F.; Ooi, B.~C.; and Miao, C., eds., \emph{{KDD} '21: The 27th {ACM} {SIGKDD} Conference on Knowledge Discovery and Data Mining, Virtual Event, Singapore, August 14-18, 2021}, 3906--3916. {ACM}.

\bibitem[{Zheng et~al.(2021)Zheng, Wang, Xu, Shen, Qin, Huai, Liu, and Chen}]{zheng2021drug}
Zheng, Z.; Wang, C.; Xu, T.; Shen, D.; Qin, P.; Huai, B.; Liu, T.; and Chen, E. 2021.
\newblock Drug Package Recommendation via Interaction-aware Graph Induction.
\newblock In Leskovec, J.; Grobelnik, M.; Najork, M.; Tang, J.; and Zia, L., eds., \emph{{WWW} '21: The Web Conference 2021, Virtual Event / Ljubljana, Slovenia, April 19-23, 2021}, 1284--1295. {ACM} / {IW3C2}.

\bibitem[{Zheng et~al.(2023)Zheng, Wang, Xu, Shen, Qin, Zhao, Huai, Wu, and Chen}]{zheng2023interaction}
Zheng, Z.; Wang, C.; Xu, T.; Shen, D.; Qin, P.; Zhao, X.; Huai, B.; Wu, X.; and Chen, E. 2023.
\newblock Interaction-aware Drug Package Recommendation via Policy Gradient.
\newblock \emph{{ACM} Trans. Inf. Syst.}, 41(1): 3:1--3:32.

\end{thebibliography}

\end{document}